\documentclass[10pt,journal,compsoc]{IEEEtran}

\pdfoutput=1
\usepackage{graphicx}%
\usepackage{multirow}%
\usepackage{amsmath,amssymb,amsfonts}%
\usepackage{amsthm}%
\usepackage{mathrsfs}%
\usepackage{xcolor}%
\usepackage{textcomp}%
\usepackage{manyfoot}%
\usepackage{booktabs}%
\usepackage{algorithm}%
\usepackage{algorithmicx}%
\usepackage{algpseudocode}%
\usepackage{listings}%
\usepackage{tabularx} 
\usepackage{array} 
\newcolumntype{C}[1]{>{\centering\arraybackslash}m{#1}}
\usepackage{hyperref}
\usepackage{wrapfig}
\usepackage{threeparttable}

\ifCLASSOPTIONcompsoc
  \usepackage[nocompress]{cite}
\else
  \usepackage{cite}
\fi
\ifCLASSINFOpdf
\else
\fi
\begin{document}
%
\title{HQViT: Hybrid Quantum Vision Transformer for Image Classification}
%
%
%
%

\author{Hui~Zhang,
        Qinglin~Zhao*,~\IEEEmembership{Senior Member,~IEEE,}
        Mengchu~Zhou*,~\IEEEmembership{Fellow,~IEEE,}
        and~Li~Feng
\IEEEcompsocitemizethanks{\IEEEcompsocthanksitem H. Zhang, Q. Zhao (corresponding author) and L. Feng are with Faculty of Innovation Engineering, Macau University of Science and Technology, Macao 999078, China. \protect\\
E-mail: h.zhang2023@hotmail.com; qlzhao@must.edu.mo; lfeng@must.edu.mo
\IEEEcompsocthanksitem M. Zhou (co-corresponding author) is with the Department of Electrical and Computer Engineering, New Jersey Institute of Technology, Newark, NJ 07102 USA
E-mail: zhou@njit.edu }
\thanks{}}

%
%

\markboth{}%
{Shell \MakeLowercase{\textit{et al.}}: HQViT: Hybrid Quantum Vision Transformer for Image Classification}
%



\IEEEtitleabstractindextext{%
\begin{abstract}
Transformer-based architectures have revolutionized the landscape of deep learning. In computer vision domain, Vision Transformer demonstrates remarkable performance on par with or even surpassing that of convolutional neural networks. However, the quadratic computational complexity of its self-attention mechanism poses challenges for classical computing, making model training with high-dimensional input data, e.g., images, particularly expensive. To address such limitations, we propose a Hybrid Quantum Vision Transformer (HQViT), that leverages the principles of quantum computing to accelerate model training while enhancing model performance. 
HQViT introduces whole-image processing with amplitude encoding to better preserve global image information without additional positional encoding.
By leveraging quantum computation on the most critical steps and selectively handling other components in a classical way, we lower the cost of quantum resources for HQViT. The qubit requirement is minimized to $O(log_2N)$ and the number of parameterized quantum gates is only $O(log_2d)$, making it well-suited for Noisy Intermediate-Scale Quantum devices.
By offloading the computationally intensive attention coefficient matrix calculation to the quantum framework, HQViT reduces the classical computational load by $O(T^2d)$. 
Extensive experiments across various computer vision datasets demonstrate that HQViT outperforms existing models, achieving a maximum improvement of up to $10.9\%$ (on the MNIST 10-classification task) over the state of the art. This work highlights the great potential to combine quantum and classical computing to cope with complex image classification tasks.
\end{abstract}

}

\maketitle

\IEEEdisplaynontitleabstractindextext

%
\IEEEpeerreviewmaketitle

\IEEEraisesectionheading{\section{Introduction}\label{sec:introduction}}

%
%
%
%

\subsection{Background}\label{sec1.1}

\IEEEPARstart{I}{N} 2017, the emergence of Transformer \cite{bib28} revolutionized the landscape of deep learning. Initially, it demonstrated its strength in the field of natural language processing (NLP), quickly surpassing both RNN and LSTM architectures to become the dominant model in NLP \cite{bib46, bib47}. Vision Transformer (ViT) was then proposed \cite{bib29}, extending the transformer architecture to the computer vision (CV) domain. Once again, the Transformer showcased its remarkable performance, achieving results on par with, or even exceeding, those of convolutional neural networks (CNNs). As a result, Transformer is now applied across various CV tasks, e.g., image restoration, image generation, and image segmentation \cite{bib48, bib49, bib56}.
At the heart of the transformer architecture lies self-attention mechanism, enabling the model to weigh the importance of different parts of input data, effectively capturing long-range dependencies. This mechanism is the key to the Transformer’s outstanding learning capabilities. However, its related computational complexity grows quadratically with the sequence length, significantly increasing the computational demands. This makes the training of high-dimensional input data extremely resource-intensive for classical computing systems.

Conversely, quantum machine learning (QML) has emerged rapidly in recent years, combining the strengths of quantum computing with classical machine learning techniques \cite{bib1, bib8}. QML aims to leverage quantum phenomena like superposition and entanglement as new computing resources to accelerate machine learning tasks \cite{bib7, bib9, bib12}. Motivated by this potential, many quantum counterparts to classical neural network models have been developed, such as QCNNs \cite{bib39, bib18, bib19, bib20, bib54}, QRNNs \cite{bib22, bib23}, and QGANs \cite{bib24, bib25}. They are based on a paradigm known as a Variational Quantum Algorithm (VQA) \cite{bib27}, which involves replacing certain components of classical neural networks with parameterized quantum circuits and using classical optimizers to update these parameters \cite{bib4}.

Given the importance of image processing in machine learning and the significant impact of the transformer architecture on computer vision, a promising idea is to propose a quantum analog of Transformer. This approach aims to leverage the advantages of quantum computing to accelerate current CV tasks. Yet research on quantum Transformer models, especially for vision tasks, is in its early stage, with only a handful of studies currently in progress. Most existing work focuses on text-based tasks, and there are even fewer studies dedicated to Quantum Vision Transformers. Although some quantum Transformer models targeted for NLP can be adapted for CV tasks, they are limited in the image size they can handle or require the pre-extraction of image features. Another issue is that the current methods are not well-optimized in balancing quantum resources and classical computation acceleration. Some approaches only quantumize a specific component of the transformer \cite{bib30, bib40}, which, while requiring fewer quantum resources, do not provide substantial acceleration to the model training. On the other hand, some methods that adopt a fully quantum approach \cite{bib31, bib41, bib42}, can drastically reduce computational loads, but they consume substantial quantum resources, making them difficult to be scalable on Noisy Intermediate-Scale Quantum (NISQ) era, which refers to an early stage of quantum computers. NISQ devices have a limited number of qubits, and these devices are affected by noise and the limited coherence time of qubits, which prevents the quantum circuits from being designed too deeply.

\subsection{Motivation}\label{sec1.2}

\begin{figure}[t]
    \centering
    \includegraphics[width=0.5\textwidth]{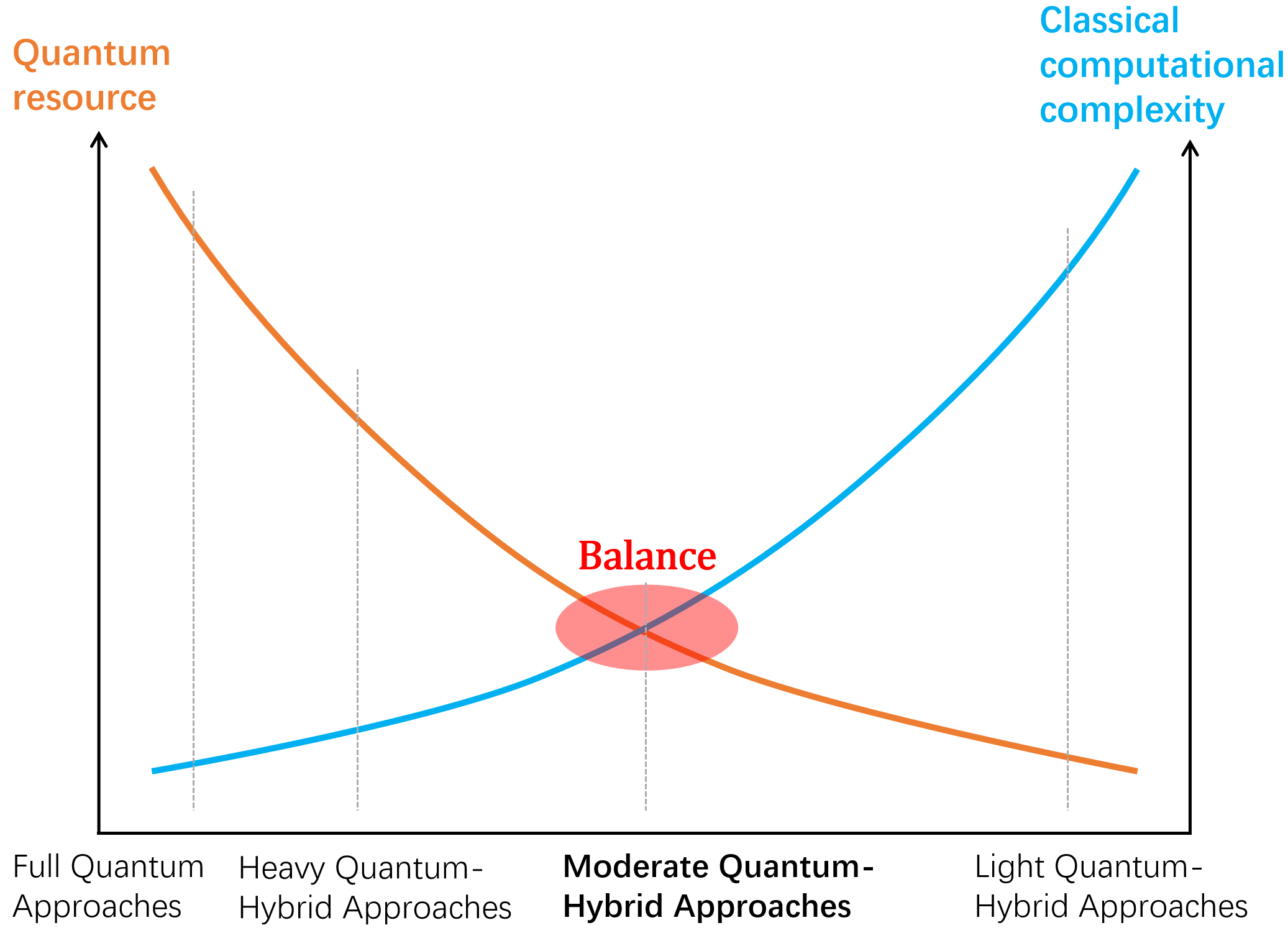}
    \caption{The trade-off between the amount of quantum resource and classical computational complexity for existing quantum self-attention and quantum transformer models. Our motivation is to achieve a balance between the two, ensuring that the quantum resources align with the limitations of NISQ devices while maintaining high-performance and scalability. The vertical axis on the left represents the quantum resource requirements, while the vertical axis on the right represents the classical computational complexity of the model. Both axes are dimensionless, since we do not conduct a quantitative comparison between the two, merely illustrating the relative relationships of the quantum resource requirements (or classical computational complexity) across different models. }\label{balance}
\end{figure}

Therefore, we can conclude that there is a clear need to develop Quantum Vision Transformer model specifically tailored for image processing. As illustrated in Figure \ref{balance}, excessive or insufficient quantum hybridization can lead to an imbalance between the model's quantum resource requirements and classical complexity. Therefore, a moderate quantum-classical hybrid approach appears to be a promising pathway, striking a balance between reducing classical computational load and efficiently using quantum resources. Such a design would not only enhance the practicality of the model for NISQ systems but also ensure superior performance in CV tasks.

\subsection{Contributions}\label{sec1.3}

\begin{figure}[h]
    \centering
    \includegraphics[width=0.5\textwidth]{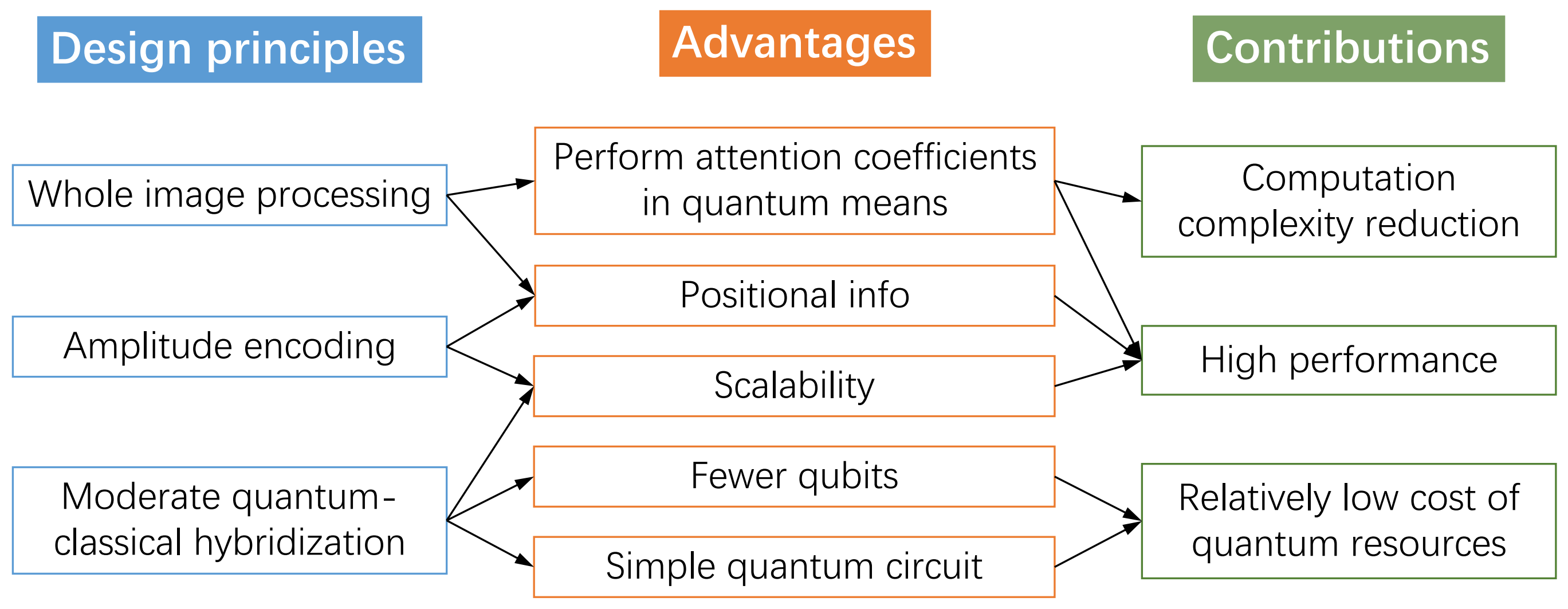}
    \caption{Overview of HQViT's design principles, advantages, and contributions.}\label{fig-contribution}
\end{figure}

To address this challenge, we propose a Hybrid Quantum Vision Transformer (HQViT) that incorporates whole-image processing and amplitude encoding. By capturing the similarity between $Q$ and $K$ within a quantum system, our model computes attention coefficients in the quantum framework, significantly reducing computational complexity. Moreover, by moderately integrating quantum and classical components, our model balances the reduction of classical computational demands and the requirements of quantum resources. The design principles, advantages and contributions of our approach are shown in Fig. \ref{fig-contribution}, with the specific descriptions as follows:

1) Whole-Image Processing. We process the image data as a whole rather than handling patch-by-patch separately as in conventional Transformer \cite{bib28}. This enables the model to capture the relationships among tokens within the quantum framework, without needing to convert $Q$ and $K$ into classical data, thereby reducing classical computation complexity by $O(T^2d)$, where $T$ is the number of patches and $d$ is the dimension of each patch.

2) Amplitude Encoding. Our model utilizes amplitude encoding to efficiently manage high-dimensional input data without requiring a large number of qubits. This also serves as the foundation for the implementation of whole-image processing. Another beneficial byproduct of amplitude encoding combined with whole-image processing is that it simultaneously accounts for the positional information of the tokens, thereby allowing the model to be less reliant on additional positional encoding. Thus, the model architecture is further simplified.

3) Moderate quantum-classical hybrid architecture. We selectively retain some classical components because their computational complexity is relatively low, and quantumizing them would require substantial quantum resources. This approach ensures the efficiently use of quantum resources, which are then dedicated to addressing the most computationally intensive steps in the model. This reduces the requirement of qubits to $O(log_2Td)$ and the number of parameterized quantum gates to $O(log_2d)$, making it applicable for NISQ devices. Meanwhile, this moderate hybrid approach enhances scalability and maintains high performance, achieving a maximum improvement of up to $11\%$ (on the MNIST 10-classification task) over the state of the art.

4) Performance evaluation. Experimental results across multiple computer vision datasets of varying scales demonstrate that HQViT outperforms existing models in most classification tasks. It does so while maintaining a limited of quantum resources and relatively low classical computational complexity.

The remainder of the paper is organized as follows: Section 2 presents related work, Section 3 introduces some quantum computing concepts related to our model, Section 4 describes the methodology and model architecture, Section 5 presents the experimental implementation and results, and Section 6 concludes the study.

\section{Related Work}\label{sec2}

Several approaches to implementing quantum-enhanced self-attention mechanisms have emerged in recent years, each differing in the level of quantum integration. These approaches can be broadly categorized into "light quantum-hybrid", "heavy quantum-hybrid", and "full quantum" models.

\textbf{Light Quantum-Hybrid Approach.} In 2022, Li et al. \cite{bib30} proposed a Quantum Self-Attention Neural Network (QSANN), which replaced classical linear mapping matrices ($W_q$, $W_k$, and $W_v$) with three parameterized quantum circuits (PQCs). These PQCs mapped the input data into $Q$, $K$, and $V$, followed by classical computation of attention coefficients. The motivation behind using PQCs was to project classical data into quantum-enhanced feature space, with the goal of extracting features that are difficult for classical linear mappings to capture. QSANN achieved better performance than that of its classical counterpart, demonstrating the powerful capabilities of quantum computing. Zhang et al. \cite{bib40} extended this idea with an improved Quantum Self-Attention Model (QSAM). By introducing amplitude-phase decomposed measurements and more expressive PQCs, QSAM extracted quantum state information more efficiently and reduced the number of learnable parameters by one-third. However, both QSANN and QSAM require a) converting $Q$, $K$, and $V$ from quantum states back to classical states and b) computing the attention coefficients classically, thus limiting the reduction of classical computational complexity.

\textbf{Heavy Quantum-Hybrid Approach.} In 2024, El Cherrat et al. introduced the Quantum Orthogonal Vision Transformer (QViT) \cite{bib42}, which computed the attention coefficient matrix and weighted Values entirely through quantum means. The remaining classification task was completed by a classical fully-connected layer. The model's backbone was a quantum orthogonal neural network built by using Reconfigurable Beam Splitter (RBS) gates \cite{bib61}. A dedicated data loader allowed the quantum circuit to load the entire sequence, capture relationships among tokens, and computed weighted values directly within the quantum framework. This method achieved comparable classification accuracy on the MedMNIST dataset against a classical baseline. However, the RBS gate are not a hardware-efficient gate and loading the entire image lead to an extremely deep circuit. hindering its practical deployment on NISQ devices. Another heavy quantum-hybrid quantum method is given by Zhao et al. \cite{bib62}, which utilizes Quantum Kernel to capture the similarity between $Q$ and $K$, and then associates the similarity information to $V$ by deferred conditional measurements. For quantum kernel methods to be effective, the measurement outcome of the output quantum state must match the initial state, which requires discarding a large number of measurement results, significantly increasing the sampling cost.

\textbf{Full Quantum Approach.} Shi et al. developed an end-to-end Quantum Self-Attention Network (QSAN) in 2024 \cite{bib31}. They introduced a novel quantum interpretation of self-attention mechanism called Quantum Logical Similarity (QLS), which utilizes quantum logical operations to capture the similarity among tokens, allowing uninterrupted execution of the model on quantum computers. However, QLS required a $O(T^2)$ complexity on number of qubits, making the model hard to be scalable for NISQ devices. Another ``Full-Quantum” Self-Attention Neutral Network (F-QSANN) was proposed by Zheng et al. \cite{bib41} in 2023. This approach entangled the elements of $Q$, $K$, and $V$ by using multi-control quantum gates, and then measured the qubits in the $V$ register to obtain the weighted values, which carry certain relational information between $Q$ and $K$. However, since the similarity relationships are captured along the depth of the circuit, the complexity of circuit depth of this method is $O(T^2)$. So it is also very challenging to make this method practical on NISQ devices.

From the above overview, it is evident that ``light quantum” approaches are insufficient in reducing classical computational complexity, while ``heavy quantum” or ``full-quantum” models suffer from poor scalability in terms of both input size and circuit depth, Making them difficult to scale to practical image tasks. Therefore, a ``moderate quantum-classical hybrid” approach appears to be a more promising solution for practical quantum computing in the NISQ era. Balancing the consumption of quantum resources and the reduction of classical computational load is key to designing effective quantum transformer models in the computer vision (CV) domain.

To address these challenges, this work proposes a novel Hybrid Quantum Vision Transformer (HQViT). It can well leverage a quantum neural network to handle image data without requirement of preprocessing. Based on the ``moderate quantum-classical hybrid” concept, HQViT leverages a quantum self-attention module to handle the most computationally intensive step in Transformer, i.e., the computation of attention coefficients matrix, and selectively retains several classical modules, making it feasible for deployment on NISQ devices. Experimental results indicate that our model achieves comparable or even better performance than existing quantum self-attention and transformer models, showcasing its effectiveness and practicality in real-world applications.


\section{Preliminaries}\label{sec3}

Before delving into our quantum vision transformer model, it is necessary to understand a few fundamental concepts of quantum mechanics, including quantum states, ansatz, swap test and measurement.

\textbf{Quantum States.} In quantum computing, quantum information is usually represented by $n$-qubit (pure) quantum states over Hilbert space $\mathbb{C}^{2^n}$. A quantum state is typically represented by using Dirac notation, such as \(|\psi\rangle\). For a single qubit, the state can be written as:
\begin{equation}  
|\psi\rangle = \alpha|0\rangle + \beta|1\rangle
\end{equation}  
where $|0\rangle$ and $|1\rangle$ are the basis of Hilbert space, and $\alpha$, $\beta$ are amplitudes, which are complex numbers satisfying \(|\alpha|^2 + |\beta|^2 = 1\). The values of $\alpha$ and $\beta$ describe the probability distribution of the qubit being in the $|0\rangle$ or $|1\rangle$ state.

\textbf{Ansatz.} An ansatz is a proposed form for a quantum state or a quantum circuit, often used as an initial guess in variational algorithms. It is a parameterized quantum circuit designed to approximate the desired quantum state or solve a particular problem. It contains a set of parameterized quantum gates, and the rotation angles of these gates can be adjusted during the training process. In the VQA framework, the parameters in an ansatz are typically optimized by using a classical optimizer \cite{bib60}. For example, an ansatz operated on a two-qubit system can be written as:
\begin{equation}
|\psi(\theta_1, \theta_2)\rangle = U(\theta_1, \theta_2)|00\rangle
\end{equation}
where $U(\theta_1, \theta_2)$ is a unitary operator parameterized by $\theta_1$ and $\theta_2$.

\begin{figure}[h]
    \centering
    \includegraphics[width=0.3\textwidth]{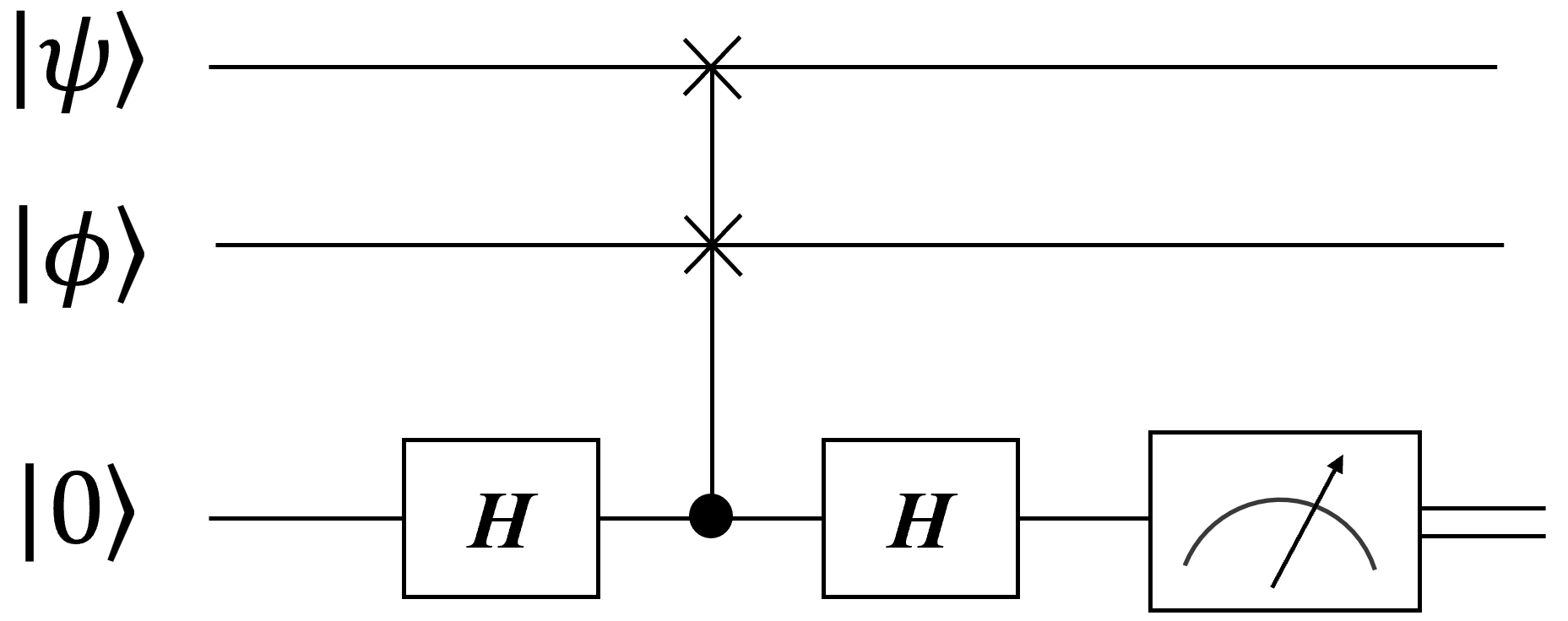}
    \caption{The swap test circuit.}\label{fig-swaptest}
\end{figure}

\textbf{Swap Test.} A swap test is a quantum operation used to determine the similarity between two quantum states \(|\psi\rangle\) and \(|\phi\rangle\). It involves applying a swap operation on the two quantum states, while the swap operation is controlled by an ancilla qubit. On the ancilla qubit, a Hadamard gate is applied before and after the controlled swap operation. The ancilla qubit is then measured, and the probability of the measurement yielding $0$ is:
\begin{equation}
Pr(0) = \frac{1 + |\langle \psi | \phi \rangle|^2}{2}.
\end{equation}
This probability reflects the degree of overlap between the two states, serving as a measure of similarity. It provides an efficient means of comparing two quantum vectors. The swap test circuit is shown in Fig. \ref{fig-swaptest}.

\begin{figure*}[h]
    \centering
    \includegraphics[width=0.85\textwidth]{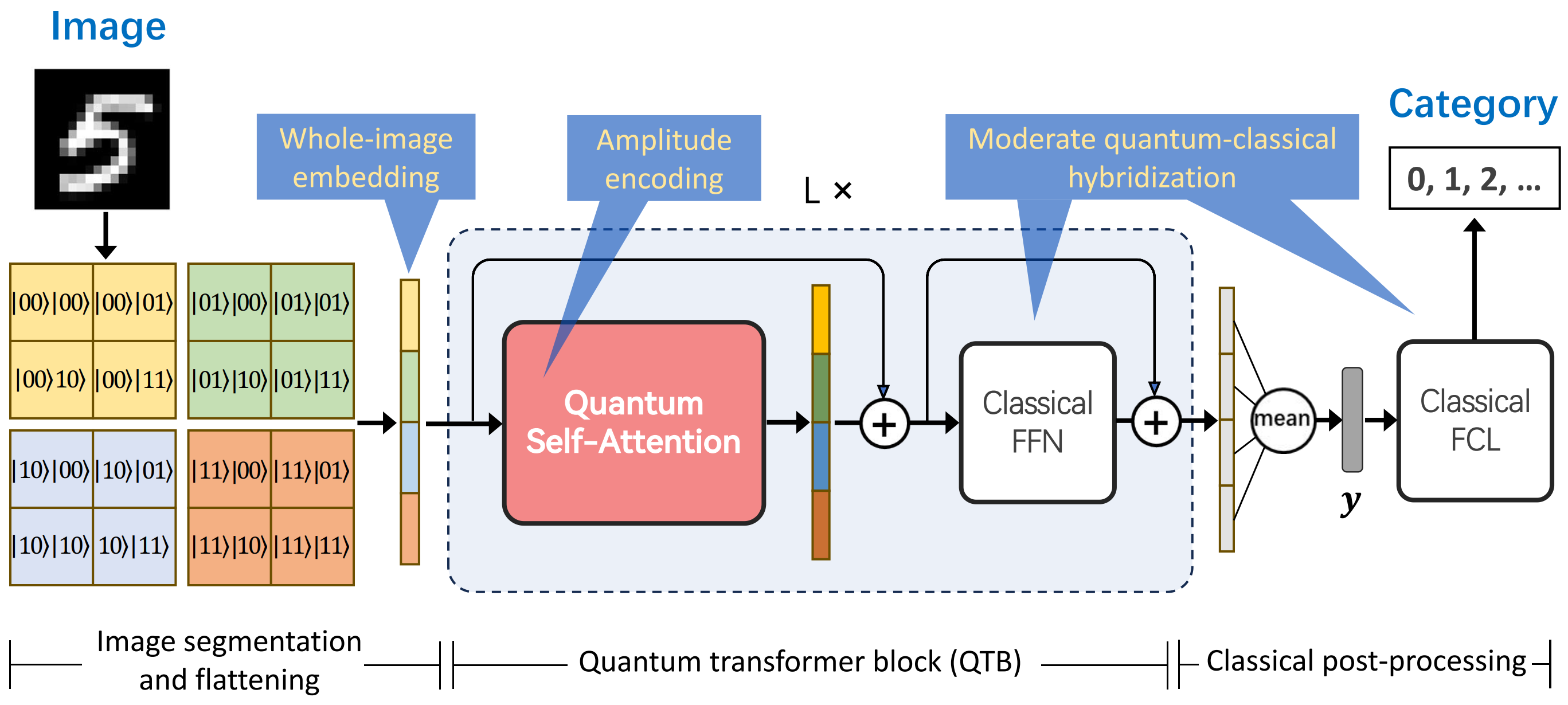}
    \caption{Overall framework of HQViT. The image is first segmented and flattened as a whole-image embedding (we use a 4x4 grayscale image as an example), which is processed by the quantum self-attention module. The output is then passed through classical feedforward layers (FFN) and further refined by using a classical fully connected layer (FCL) for final classification. The quantumized self-attention module is the core of the model, while other components are handled classically.}\label{Fig-overall framework}
\end{figure*}

\textbf{Measurement.} Quantum measurements are described by a collection $\{M_m\}$ of measurement operators. These are operators acting on the state space of the system being measured. The index $m$ refers to the measurement outcomes that may occur in the experiment. If the state of the quantum system is $|\psi\rangle$ immediately before the measurement then the probability that result $m$ occurs is
\begin{equation}
Pr(m) = \langle\psi|M_m^\dagger M_m|\psi\rangle.  \label{eq3-meassure}
\end{equation}
By selecting different projection operators, we can obtain the probability information of the quantum state projected onto different subspaces. In our model, this probability information from the different subspaces is used to obtain the attention coefficients for different token pairs.

The state of the system after the measurement is
\begin{equation}
|\psi_{\text{post}}\rangle = \frac{M_m|\psi\rangle}{\sqrt{\langle \psi |M_m| \psi \rangle}}.
\end{equation}
This describes the change in the quantum state following measurement, where the system collapses into one of the possible measured states.


\section{Hybrid Quantum Vision Transformer}\label{sec4}
\subsection{Overall architecture}\label{sec4.1}

The overall framework of the proposed hybrid quantum vision transformer (HQViT) is as shown in Fig. \ref{Fig-overall framework}, showing the integration of both quantum and classical components. Aside from the quantum self-attention module, the other classical parts largely maintain the structure of the classical Vision Transformer (ViT). First, the input image is segmented into smaller patches. Each patch is then flattened into a vector and all patch vectors are concatenated to form a single representation for the entire image, which is called whole-image embedding. This whole-image embedding is then fed into a quantum transformer block (QTB), which consists of a quantum self-attention module and a classical feed-forward neural network (FFN), both with residual connections. The quantum self-attention module, the core of the model, uses amplitude encoding and PQCs to process the self-attention mechanism and produce a set of weighted values as output. A total of $L$ QTBs are stacked to enhance the model's feature extraction capabilities. Finally, a classical post-processing step averages the weighted values from the last QTB and feeds them into a fully-connected layer (FCL) to output the predicted category. By integrating such techniques as whole image embedding and amplitude encoding, we reduce the quantum resources required for the quantum components of the model. Additionally, we selectively retain certain classical components. Hence, the overall framework showcases our "moderate quantum-classical hybridization" design principle.

For the image segmentation, similar to ViT, an image with dimensions \( H \times W \times C \) is divided into patches of size \( d = h \times w \times C \), resulting in \( T = \frac{H}{h} \times \frac{W}{w} \) patches. Here, we use a \( 4 \times 4 \) grayscale image as an example. It can be divided into 4 patches of size \( 2 \times 2 \). By Using amplitude encoding in quantum self-attention module, the image can be represented by a 4-qubit quantum state. As shown in Fig. \ref{Fig-overall framework}, each pixel's position is represented by each basis of the quantum state, where the first two qubits precisely encode the index of each patch. This index plays a crucial role in the subsequent quantum-based processing of attention coefficients, to be explained further.

\begin{figure*}[t]
    \centering
    \includegraphics[width=0.95\textwidth]{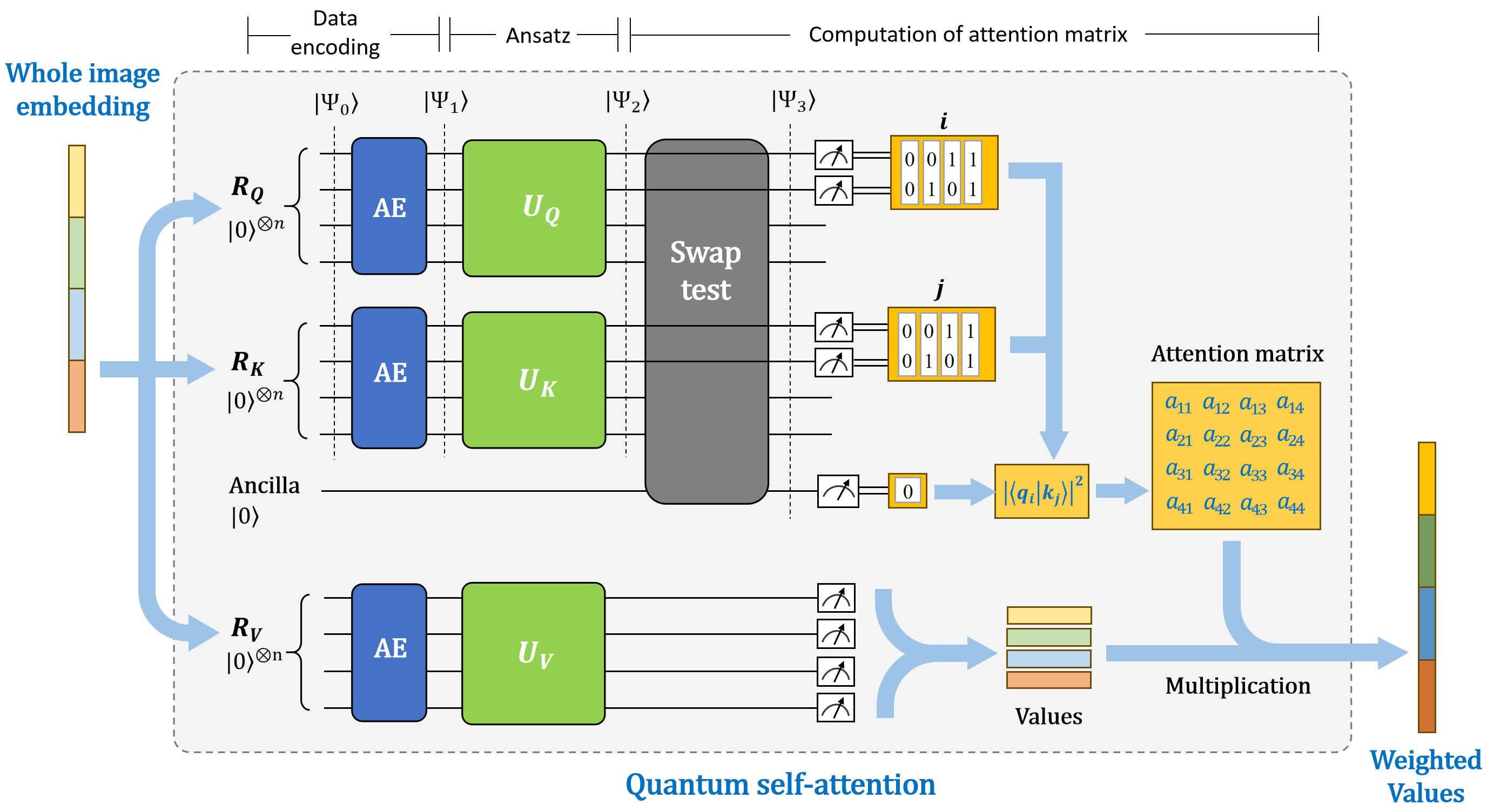}
    \caption{Quantum self-attention mechanism for HQViT. The whole-image embedding is fed into registers \(R_Q\), \(R_K\), and \(R_V\) with amplitude encoding (AE), followed by unitary transformations (\(U_Q\), \(U_K\), \(U_V\)). A swap test is performed on the \(Q\) and \(K\) quantum states to compute the attention matrix, which is then combined with the \(V\) to produce Weighted Values.}\label{Fig-quantum self-attention}
\end{figure*}

\subsection{Quantum self-attention module}\label{subsubsec4.2}

The quantum self-attention module processes the entire image within the quantum system, allowing it to capture long-range dependencies among tokens and directly compute the attention coefficient matrix. Our encoding strategy encodes both the pixel values and positions, meaning the positional information of each patch is integrated into the whole-image embedding. Fig. \ref{Fig-quantum self-attention} illustrates the structure of the quantum self-attention module. Below, we summarize the basic mechanism. For simplicity, assume that the input image $X$ consists of a total of $N$ pixels, and it is divided into $T$ patches, each with a dimension of $d$.

i) We utilize three quantum registers, denoted as $R_Q$, $R_K$, and $R_V$, to receive the input data. Each of them is initialized with the state of $|0\rangle^{\otimes n}$, where the number of qubits $n = log_2N$. We are first concerned with $R_Q$ and $R_K$. The whole image is fed into $R_Q$ and $R_K$ as a vector with amplitude encoding, obtaining $|\Psi_1\rangle$.

\begin{equation}
|\Psi_0\rangle = |0\rangle^{\otimes n} \otimes |0\rangle^{\otimes n} \xrightarrow{\text{Data encoding}} |\Psi_1\rangle = |\psi_X\rangle \otimes |\psi_X\rangle \label{eq1}
\end{equation}

ii) The encoded quantum states undergo ansatzs $U_Q$ and $U_K$, which act as the quantum equivalents of the classical linear mapping matrices $W_q$ and $W_k$. This results in quantum states $|\psi_Q\rangle$, $|\psi_K\rangle$, respectively.

\begin{equation}
|\Psi_2\rangle = (U_Q|\psi_X\rangle) \otimes (U_K|\psi_X\rangle) = |\psi_Q\rangle \otimes |\psi_K\rangle \label{eq2}
\end{equation}

iii) We then combine the registers $R_Q$, $R_K$, and an ancilla qubit to perform a swap test. Thanks to our encoding strategy, the qubits in \(R_Q\) or \(R_K\) can be considered into two parts: the index subsystem encodes the index information of the tokens and the patch subsystem holds the information of each token. The swap test is applied only to the patch subsystems of $R_Q$ and $R_K$ to extract the similarities between each pair of tokens, $q_i$ and $k_j$.

\begin{equation}
|\Psi_2\rangle \otimes |0\rangle \quad \xrightarrow{\text{Swap test}} \quad |\Psi_3\rangle \label{eq3}
\end{equation}

iv) We measure the index subsystems of $R_Q$ and $R_K$ and ancilla qubit. We keep the outcome of ancilla qubit being 0, and traverse all the outcomes of the index subsystems of \( R_Q \) and \( R_K \). This gives us a set of probabilities ${Pr(i,j,0)}_{i,j=0}^{T}$, where $i$ and $j$ are measurement outcomes of the index subsystems of \( R_Q \) and \( R_K \), respectively. Each probability corresponds to the similarity of a particular pair of $q_i$ and $k_j$. This can be expressed as:

\begin{align} 
& |\Psi_3\rangle \quad \xrightarrow{\text{Measurement}} \quad \text{outcome:  } \{i,j,0\} \\
& Pr(i,j,0) \quad \Rightarrow \quad \langle q_i | k_j \rangle
\end{align}
where \( f \) is a simple functional relationship.

v) For $R_V$, the input data is similarly encoded and fed into ansatz $U_V$. Then measurements are performed to obtain classical $V$. Finally, the attention coefficient matrix, after applying softmax, is combined with $V$ to obtain the outputs of quantum self-attention, also known as the Weighted Values.

\subsubsection{Data encoding}\label{subsec4.2.1}

The data encoding strategy is crucial in quantum neural networks (QNNs) and can even have a decisive impact on the network's design \cite{bib34}. For an input vector $\mathbf{X}=\{x_i\}_{i=0}^{N-1}$,
the amplitude encoding can be written as

\begin{equation}
|\psi\rangle = \frac{1}{\|\mathbf{X}\|}\sum_{i=0}^{N-1} x_i |i\rangle, \label{eq4.1.1-1}
\end{equation}
where $|i\rangle$ denotes the computational basis, and $\|\mathbf{X}\|$ is normalization factor. The number of qubits required for this encoding scheme is $n$, which satisfies \(2^{n-1} < N \leq 2^n\), with the amplitudes of unused basis states padded with zeros. Therefore, the required number of qubits is $O(log_2N)$.

For a whole-image embedding with $T$ patches, each of size $d$, amplitude encoding is written as:

\begin{equation}
|\psi\rangle = \sum_{i=0}^{T-1}\sum_{l=0}^{d-1}x_{il}|i\rangle|l\rangle,   \label{eq4.1.1-2}
\end{equation}
where $i$ represents the patch index, giving each patch its positional information, and $j$ represent the index of elements within each patch. From this perspective, the encoded quantum system can be seen as comprising two parts: the index subsystem (higher-order qubits), which carrys positional information, and the patch subsystem (lower-order qubits), which encodes the patch data. These subsystems play distinct roles during the subsequent evolution.

\subsubsection{Ansatz}\label{subsec4.2.2} 

\begin{figure*}[t]
    \centering
    \includegraphics[width=0.90\textwidth]{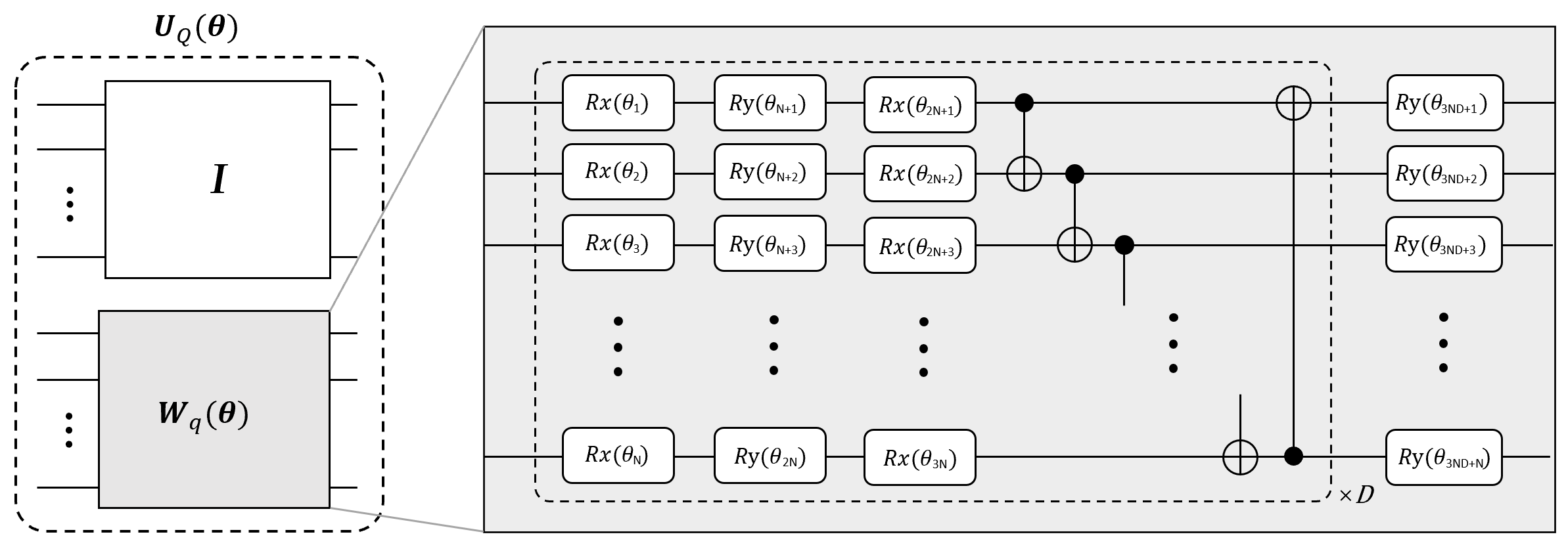}
    \caption{The structure of $U_Q$, where $n$ represents the number of qubit of $R_Q$ and $D$ denotes the number repeated basic units.}  \label{fig-ansatz}
\end{figure*}

The function of \( U_Q \), \( U_K \) and \( U_V \) is to apply linear mappings on the input data to generate \( Q \), \( K \) and \( V \). In classical self-attention, matrices \( W_q \), \( W_k \) and \( W_v \) map each token individually. In our quantum self-attention mechanism, the input is the whole image. Thus, $U_Q$ and $U_K$ are constructed to fulfill this linear mapping, i.e.,

\begin{equation}
    U_{Q} = \begin{pmatrix}
     W_{q} &  &  &  &  \\
     & W_{q} &  &  &  \\
     &  & ... &  &  \\
     &  &  & W_{q} &  \\
     &  &  &  & W_{q} 
    \end{pmatrix}
    =\begin{pmatrix}
    1 &  &  &  &  \\
     & 1 &  &  &  \\
     &  & ... &  &  \\
     &  &  & 1 &  \\
     &  &  &  & 1 
    \end{pmatrix} \otimes W_{q} .  \label{eq4.1.2-1}
\end{equation}
Here we discuss only \( U_Q \) as an example, since the principles of \( U_K \) and \( U_V \) are completely the same as \( U_Q \)'s.

In Eq. \ref{eq4.1.2-1}, $W_{q}$ is a linear mapping matrix with size of $d\times d$. So the size of $U_{Q}$ is $Td\times Td$. The evaluation of Eq.(\ref{eq2}) can be written as

\begin{equation}
\begin{aligned}
    U_{Q} |\psi_X\rangle &= \begin{pmatrix}
    W_{q} &  &  &  &  \\
     & W_{q} &  &  &  \\
     &  & ... &  &  \\
     &  &  & W_{q} &  \\
     &  &  &  & W_{q} 
    \end{pmatrix} 
    \begin{pmatrix} x_1\\
     x_2\\
     ... \\
     x_{T-1}\\
     x_T\end{pmatrix} = 
     \begin{pmatrix} q_1 \\
     q_2 \\
     ... \\
     q_{T-1} \\
     q_T \end{pmatrix}\\ &= |\psi_{Q}\rangle.  \label{eq4.1.2-2}
\end{aligned}    
\end{equation}

The circuit implementation of \( U_{Q} \) is shown in Fig. \ref{fig-ansatz}. Here, $\mathbf{I}$ represents the identity transformation, and \( W_{q}(\theta) \) is a parameterized quantum circuit, which consists of \( D \) layers of repeatable basic units. Each basic unit contains a set of single-qubit gates and a set of two-qubit gates. This structure allows the circuit to span the entire unitary group, ensuring sufficient expressibility of the model \cite{bib36}. Finally, a set of \( Ry \) gates is used for fine-tuning before outputting the quantum state.

After the linear mapping, the quantum states $|\psi_{q}\rangle$ and $|\psi_{k}\rangle$ can be written as

\begin{equation}
|\psi_Q\rangle = \sum_{i=0}^{T-1} \sum_{l=0}^{d-1} \alpha_{il} |i\rangle |l\rangle,
\quad
|\psi_K\rangle = \sum_{j=0}^{T-1} \sum_{k=0}^{d-1} \beta_{jk} |j\rangle |k\rangle.  \label{eq4.1.2-3}
\end{equation}

\subsubsection{Computation of attention matrix}\label{subsec4.2.3}

Then, we combine registers $R_Q$ and $R_K$ with an ancilla qubit and perform a swap test on the combined quantum system. After the swap test, the state of the combined quantum system becomes:

\begin{equation}
|\Psi_3\rangle = \frac{1}{2} \Big(\big(|\Psi_2\rangle + |\Psi_{sw}\rangle\big) \otimes |0\rangle + \big(|\Psi_2\rangle - |\Psi_{sw}\rangle\big) \otimes |1\rangle \Big),  \label{eq4.4-1}
\end{equation}
where $|\Psi_{sw}\rangle$ is the state resulting from applying the swap operation to $|\Psi_2\rangle$. Recall that

\begin{equation}
|\Psi_2\rangle = |\psi_Q\rangle \otimes |\psi_K\rangle = \sum_{i,l=0}^{T-1} \sum_{j,k=0}^{d-1} \alpha_{il} \beta_{jk} |i\rangle |l\rangle |j\rangle |k\rangle.   \label{eq4.4-2}
\end{equation}
Since the swap operation only involves the patch subsystems, $|\Psi_{sw}\rangle$ can be written as

\begin{equation}
|\Psi_{sw}\rangle = \sum_{i,l=0}^{T-1} \sum_{j,k=0}^{d-1} \alpha_{ik} \beta_{jl} |i\rangle |l\rangle |j\rangle |k\rangle.   \label{eq4.4-3}
\end{equation}
We notice that if we perform projection measurement on the ancilla qubit, the probability of obtaining outcome $0$ is given as:

\begin{equation}
\begin{aligned}
Pr(0)_{anc} &= \frac{1}{2} ( 1 + \langle \Psi_2|\Psi_{sw}\rangle ) \\
&= \frac{1}{2} ( 1 + \sum_{i=0}^{T-1} \sum_{j=0}^{T-1} \sum_{l=0}^{d-1} |\alpha_{il}|^2 |\beta_{jl}|^2 ) \\
&= \frac{1}{2} ( 1 + \sum_{i=0}^{T-1} \sum_{j=0}^{T-1} |\langle q_i|k_j \rangle|^2 ). \label{eq-P(0)}
\end{aligned}
\end{equation}

This probability gives us information about the overlaps (or similarities) between $|q_i\rangle$ and $|k_j\rangle$, $\forall i, j\in \{0, ..., T-1\}$. Therefore, to extract the value of each pair of $\langle q_i |k_j \rangle$, we need to measure the ancilla qubit and all qubits in the index subsystems of $R_Q$ and $R_K$. For a set of measurement operators corresponding to outcomes $\{i,j,0\}_{i,j=0}^{T-1}$, where $i$ and $j$ represent the outcome of index subsystems of $R_Q$ and $R_K$ (in the form of binary string), and $0$ is the outcome of ancilla qubit, we obtain the set of probability values $\{Pr(i,j,0)\}_{i,j=0}^{T-1}$, where each $Pr(i,j,0)$ is given:

\begin{equation}
Pr(i,j,0) = \frac{1}{2}(\|q_i\|^2\|k_j\|^2 + |\langle q_i|k_j\rangle|^2). \label{eq-measurement}
\end{equation}
Since $q_i = W_q \mathbf{x}_i$, $k_j = W_k \mathbf{x}_j$, and both $W_q$ and $W_k$ are unitary transformations, we have $\|q_i\| = \frac{1}{\|X\|}\|\mathbf{x}_i\|$ and $\|k_j\| = \frac{1}{\|X\|}\|\mathbf{x}_j\|$. So finally we obtain:

\begin{equation}
Pr(i,j,0) = \frac{1}{2}(\frac{1}{\|X\|^4}|\mathbf{x}_i|^2|\mathbf{x}_j|^2 + |\langle q_i|k_j\rangle|^2). \label{eq-measurement-2}
\end{equation}
Note that $Pr(0)_{anc} = \sum_{i,j=0}^{T-1}Pr(i,j,0)$. The detailed derivations of Eq. (\ref{eq-P(0)}) and (\ref{eq-measurement}) can be found in Supplements of this paper.

In this way, by measuring the index subsystems of $R_Q$, $R_K$ and ancilla qubit all at once, we can derive the similarities between $q_i$ and $k_j$, i.e., the attention coefficient matrix. Take \( T = 4 \) and \( d = 4 \) as an example to demonstrate this process more clearly. Both \( R_Q \) and \( R_K \) have 4 qubits each, with their index subsystems and patch subsystems each consisting of 2 qubits. So the measurement outcomes are a bench of 5-bit binary strings, denoted as $B$. The first two bits of $B$ represent $i$, the following two bits represent $j$, and the last bit corresponds to the measurement outcome of ancilla qubit, where we only care about the result of 0. When \( B = 00000 \), the swap test yields $Pr(0,0,0)$, allowing us to obtain $\langle q_0 |k_0 \rangle$; when \( B = 00010 \), it yields the $Pr(0,1,0)$ and gives $\langle q_0 |k_1 \rangle$, and so on. This process yields the entire attention coefficient matrix (ACM):

\begin{equation}
\begin{aligned}
    &B = \begin{pmatrix}
    00000 & 00010 & 00100 & 00110 \\
    01000 & 01010 & 01100 & 01110 \\
    10000 & 10010 & 10100 & 10110 \\
    11000 & 11010 & 11100 & 11110 \\
    \end{pmatrix} \Rightarrow \\
    &ACM = \begin{pmatrix}
    |\langle q_0|k_0\rangle|^2 & |\langle q_0|k_1\rangle|^2 & |\langle q_0|k_2\rangle|^2 & |\langle q_0|k_3\rangle|^2 \\
    |\langle q_1|k_0\rangle|^2 & |\langle q_1|k_1\rangle|^2 & |\langle q_1|k_2\rangle|^2 & |\langle q_1|k_3\rangle|^2 \\
    |\langle q_2|k_0\rangle|^2 & |\langle q_2|k_1\rangle|^2 & |\langle q_2|k_2\rangle|^2 & |\langle q_2|k_3\rangle|^2 \\
    |\langle q_3|k_0\rangle|^2 & |\langle q_3|k_1\rangle|^2 & |\langle q_3|k_2\rangle|^2 & |\langle q_3|k_3\rangle|^2 \\
    \end{pmatrix} .
\end{aligned}
\end{equation}

\subsection{Classical FFN and post-processing}\label{subsec4.3}

After obtaining the weighted values from the quantum self-attention module, they are passed into a FFN to integrate the internal features of each token. This step can be implemented by using either quantum or classical methods. Since FFN is not a major computational bottleneck in the transformer architecture, a classical approach is preferred to save quantum resources. In this paper, we employ a classical Multi-Layer Perceptron (MLP), consisting of one hidden layer and one output layer. The dimensionality of the hidden layer is set based on the experimental scenario, while the output dimension matches the input dimension $d$.

After the last QTB, a set of outputs, $\mathbf{y}_i$, is produced. We take the mean of them to form a global feature vector of the whole image, $\mathbf{y}$. Finally, a classical fully-connected layer is applied to make the prediction: $y:=\sigma(\mathbf{w}^T\cdot\mathbf{y}+b)$, where $\sigma$ is the sigmoid function for binary classification, or softmax funtion for multi-class classification. We use cross-entropy as the loss function. Our model is based on the Variational Quantum Algorithm (VQA) paradigm, where a classical optimizer (in our case, Adam \cite{bib58}) is used to train both quantum and classical parameters.

\subsection{Quantum resources and complexity analysis}\label{subsec4.4}

\begin{table*}[h]
\centering
\begin{threeparttable}
\caption{Quantum resource needs and classical complexity reduction of related models. The letters following the model names indicate the level of quantum-classical hybridization, where L stands for light, M for moderate, H for heavy, and F for fully quantum models.}
\label{Quantum resource}
\begin{tabular}{C{2.5cm}C{2cm}C{2cm}C{2cm}C{2cm}C{2.8cm}C{1.5cm}}
\toprule%
\textbf{Model} & \textbf{\# Qubits} & \textbf{\# PQG} & \textbf{\# Distinct circuits} & \textbf{\# Total measurement} & \textbf{Outputs of quantum circuits} & \textbf{Classical complexity reduction} \\ \midrule
QSANN \cite{bib30} (L) & $O(d)$ & $O(d))$ & $O(T)$ & $O(Td)$ & $\{\textbf{q}_i\}_{i=1}^T$, $\{\textbf{k}_i\}_{i=1}^T$, $\{\textbf{v}_i\}_{i=1}^T$ & $O(Td^2)$ \\ \midrule
QSAM \cite{bib40} (L) & $O(d)$ & $O(d))$ & $O(T)$ & $O(Td)$ & $\{\textbf{q}_i\}_{i=1}^T$, $\{\textbf{k}_i\}_{i=1}^T$, $\{\textbf{v}_i\}_{i=1}^T$ & $O(Td^2)$ \\ \midrule
HQViT (M) & \textbf{O(log(Td))} & \textbf{O(logd)} & \textbf{O(1)} & $O(T^2logd)^{*}$ & $\{A_{ij}\}_{i,j=1}^T$, $\{\textbf{v}_i\}_{i=1}^T$ & $\mathbf{O(T^2d)}$ \\ \midrule
$\text{QViT}^1$ \cite{bib42} (H) & $O(T+d)$ & $O(dlogd)$ & $O(T)$ & $O(Td)^{*}$ & $\{\textbf{y}_i\}_{i=1}^T$ & $O(T^2d)$ \\ \midrule
$\text{OKSAN}^2$ \cite{bib62} (H) & $O(d)$ & $O(d))$ & $O(T^2)$ & $O(T^2d)^{*}$ & $\{\textbf{y}_i\}_{i=1}^T$ & $O(T^2d)$ \\ \midrule
QSAN \cite{bib31} (F) & $O(Tlogd+T^2)$ & $O(Tlogd)$ & $O(1)$ & $O(logd)$ & {\textbf{y}} & $O(T^2d)$ \\ \midrule
F-QSANN \cite{bib41} (F) & $O(Tlogd)$ & $O(T^2logd)$ & $O(1)$ & $O(1)$ & Probability of the predicted class & $O(T^2d)$ \\
\bottomrule
\end{tabular}
\begin{tablenotes}
    \footnotesize
    \item 1: This paper proposes several quantum transformer approaches, and we select the one that is most closely aligned with the standard classical self-attention mechanism, specifically the approach described in Section 3.3.
    \item 2: This method has been implemented using both amplitude encoding and qubit encoding, but here we select the case of qubit encoding.
    \item '*' indicates that this method may require more sampling per measurement than other methods.
\end{tablenotes}
\end{threeparttable}
\end{table*}

We analyze the quantum resource cost and reduction in classical complexity for HQViT and compare these with other quantum self-attention or quantum transformer models. From Table \ref{Quantum resource}, it can be seen that HQViT maintains relatively low or the lowest levels across multiple quantum resource metrics (e.g., the number of qubits is \( O(\log(Td)) \), which is relatively low, while the number of PQCs and distinct circuits are the lowest). At the same time, it ensures a significant reduction in classical computational complexity. Although its measurement count is relatively high, overall, it achieves a well-balanced efficiency in quantum resource utilization compared to other methods. The key insights are given as follows:

\textbf{\# qubits.} HQViT uses amplitude encoding and requires $O(log(Td))$ qubits, which is relatively low among all methods. Additionally, models with low qubit requirements include QSANN, QSAM, and QKSAN. These models use qubit encoding and process tokens one by one in the quantum circuit. As a result, the qubit complexity has a linear relationship with the token dimension, $O(d)$ (whether $O(d)$ or $O(log(Td))$ is lower depends on the specific situation, but both are significantly lower than the qubit complexity of other models). On the other hand, other models, especially fully quantum models, require much higher qubits compared to the other models.

\textbf{\# PQGs.} In shallow parameterized quantum circuits, the number of parameterized quantum gates is generally proportional to the number of qubits for encoding a single token. However, due to the special circuit structures in QSAN and F-QSANN, their PQG requirements increase significantly. In contrast, HQViT only requires $O(\log_2d)$ PQGs, offering a light-weight circuit structures.

\textbf{\# Distinct circuits.} This metric refers to the number of distinct circuits required to process all tokens in the sequence in parallel or to compute the attention matrix in parallel. For QSANN and QSAM, tokens are fed into the quantum circuit one by one, so mapping all tokens to QKV requires \( O(T) \) distinct circuits. For QKSAN, since each circuit computes a single attention score, computing \( T^2 \) attention scores requires \( T^2 \) distinct circuits. In contrast, HQViT, QSAN, and F-QSANN use global encoding, where the attention coefficient matrix or self-attention output (or the final classification result) is computed within a holistic quantum framework, requiring only one distinct circuit.

\textbf{Measurement complexity.} To estimate measurement probabilities or expectation values, practical experiments typically require \( O(1/\epsilon^2) \) samples (where \( \epsilon \) is the target accuracy). However, in this paper, for a more intuitive comparison, we assume that applying a projection operator measurement to a single qubit counts as one measurement, ignore the detailed physical implementations. For sampling complexity, we only provide qualitative annotations (*) rather than precise calculations. Based on this standard, the measurement complexity of HQViT is $O(T^2logd)$, with \( T^2 \) post-selection measurements and each measuring \( \log d \) qubits. QKSAN requires \( O(T^2 d) \) measurements, as it runs the quantum kernel circuit \( T^2 \) times, measuring \( d \) qubits each time. QSANN and QSAM require \( O(Td) \) measurements to obtain the mapped $QKV$ vectors. In contrast, QSAN and F-QSANN only require \( O(\log d) \) and \( O(1) \) measurements, respectively, since the former's quantum circuit outputs the weighted features of all tokens, while the latter directly produces a predicted classification result. Although these fully quantum models have lower measurement complexity, their high resource demands in terms of qubits and circuit depth make them less scalable.

\textbf{Classical complexity reduction.} The computational complexity of classical transformers primarily lies in the calculation of attention coefficients, with a complexity of $O(T^2d)$. By replacing the classical self-attention mechanism with a quantum self-attention counterpart, we transfer this computational burden to the quantum system, effectively eliminating this portion of the computational cost. In comparison to lightly quantumized models, which only reduce the classical computational overhead of the linear mapping step, $O(Td^2)$, HQViT eliminates a greater amount of classical computational overhead. Furthermore, the classical computational overhead reduced by HQViT can be comparable to that of heavy quantum-hybrid and fully quantum models, i.e., $O(T^2d)$, while requiring fewer quantum resources.

\begin{table*}[h]
    \centering
    \caption{Key information about the datasets used in the experiments.}
    \begin{tabular}{C{2.5cm}|C{2cm}|c|C{1.8cm}|p{6.3cm}}
        \toprule%
        \textbf{Dataset} & \textbf{Number of Classes} & \textbf{Image Size} & \textbf{Total samples} & \textbf{Description of Image Content} \\
        \midrule
        MNIST & 10 & 28x28 & 70000 & Grayscale images of handwritten digits from 0 to 9 \\
        \midrule
        MedMNIST2D & Diverse classes & 28x28 & From 100 to 100,00  & A datasets collection of biomedical images, including sub-datasets such as ChestMNIST, DermaMNIST etc.\\
        \midrule
        CIFAR-10 & 10 & 32x32 & 60000 & Color images of real-life objects, including airplanes, cars, birds, etc. \\
        \midrule
        CIFAR-100 & 100 & 32x32 & 60000 & Color images of real-life objects, with more and finer categories than CIFAR-10 \\
        \midrule
        Mini-ImageNet & 100 & 84x84 & 60000 & Color images of real-life objects, commonly used for few-shot learning \\
        \midrule
    \end{tabular}
    \label{tab1}
\end{table*}


\section{Experiments}\label{sec5}

\subsection{Experimental settings}\label{subsec5.1}

To comprehensively evaluate our model, we have conducted experiments on several datasets, including MNIST, MedMNIST2D, CIFAR-10, and Mini-ImageNet. These datasets vary in terms of classification categories, input image size, and feature complexity, providing a robust platform to assess the scalability of HQViT. The basic information for each dataset is provided in Table \ref{tab1}.

For MNIST and MedMNIST2D, we use 8 qubits to encode each image, which results in an image size of 16x16. For CIFAR-10 and Mini-ImageNet, we conduct experiments with both 8 qubits and 10 qubits to encode images with sizes 16x16 and 32x32, respectively. For the 16x16 input size, each image is segmented into 4x4 patches, yielding 16 tokens, each with a dimension of 16. For the 32x32 input size, each image is divided into 64 patches. To ensure fair comparisons, the input images for other models used in the comparison are resized to the same dimensions. We conduct these experiments by using Tencent's tensorcircuit platform \cite{bib57}. All the images used in experiments are grayscale images.

For the hyperparameters, the number of repeatable units in each PQC, denoted $D$, is set to 2 for most experiments, except for MedMNIST2D, where we set $D=3$. The hidden layer of MLP is set to the same size as the dimension of each patch, and the number of QTBs $L$ is set to 3.

\begin{table*}[h]
\centering
\caption{Test accuracy of HQViT compared with TTN/MERA, QCNN-1, QCNN-2 and QSAM on MNIST. The highest accuracy in each column is indicated in bold font. Note: '——' indicates that there is no relative experimental data.}
\begin{tabular}{ccccccc}
\toprule%
Method                  & 0 or 1     & 1 or 8     & odd or even  & $>$4 or $\leq4$  & 0 to 9  \\
\midrule
TTN/MERA \cite{bib38}   & 0.997  & ——      & 0.848  & 0.791  & ——      \\ 
QCNN-1 \cite{bib20}     & 0.985  & ——      & ——      & ——      & ——      \\ 
QCNN-2 \cite{bib19}     & ——      & 0.963  & ——      & ——      & 0.743  \\ 
QSAM \cite{bib40}       & 0.999  & 0.995  & 0.899  & 0.844  & 0.822  \\
\midrule
HQViT                  & $\mathbf{1.000}$ & $\mathbf{1.000}$  & $\mathbf{0.964}$ & $\mathbf{0.942}$  & $\mathbf{0.931}$ \\
\midrule
\end{tabular}
\label{tab-mnist}
\vspace{0.5cm}

\centering
\begin{threeparttable}
\caption{Test accuracy of HQViT compared with QViT on MedMNIST2D. The highest accuracy in each column is indicated in bold font.}
\begin{tabular}{ccccccc}
\toprule%
Method           & BloodMNIST   & DermaMNIST   & OCTMNIST     & PneumoniaMNIST   & BreastMNIST \\
\midrule
$\text{QViT}^1$ \cite{bib42} & \textbf{0.860} &  0.719     &  0.606       & \textbf{0.885}   & 0.744  \\
\midrule
HQViT           & 0.838     &  \textbf{0.725}   & \textbf{0.608}   &  0.882    & $\mathbf{0.795}$ \\
\midrule
\end{tabular}
\begin{tablenotes}
    \footnotesize
    \item 1: Here, we use the experimental results of the OrthoTransformer model proposed in that paper, which align with the model selected for complexity analysis.
\end{tablenotes}
\end{threeparttable}
\label{tab-medmnist}
\vspace{0.5cm}

\centering
\caption{Test accuracy of HQViT on CIFAR-10 compared with CViT and QSAM. The highest accuracy in each column is indicated in bold font.}
\begin{tabular}{c|ccc|ccc}
\toprule%
\multirow{2}{*}{Method} & \multicolumn{3}{c|}{Size = $16\times 16$} & \multicolumn{3}{c}{Size = $32\times 32$} \\
\cmidrule{2-7}
       & 2-class & 4-class & 10-class & 2-class & 4-class & 10-class \\
\midrule
CViT & 0.862 & 0.587 & 0.303 & 0.883 & 0.610 & \textbf{0.335} \\
QSANN \cite{bib30} & 0.825 & 0.541 & 0.295 & 0.852 & 0.585 & 0.317 \\
QSAM \cite{bib40} & 0.833 & 0.534 & 0.312 & 0.874 & 0.607 & 0.311 \\
\midrule
HQViT & \textbf{0.875} & \textbf{0.598} & \textbf{0.335} & \textbf{0.885} & \textbf{0.615} & 0.334 \\
\midrule
\end{tabular}
\label{tab-cifar10}
\vspace{0.5cm}

\centering
\caption{Test accuracy of HQViT on Mini-ImageNet compared with CViT and QSAM. The highest accuracy in each column is indicated in bold font.}
\begin{tabular}{c|ccc|ccc}
\toprule%
\multirow{2}{*}{Method} & \multicolumn{3}{c|}{Size = $16\times 16$} & \multicolumn{3}{c}{Size = $32\times 32$} \\
\cmidrule{2-7}
       & 2-class & 4-class & 10-class & 2-class & 4-class & 10-class \\
\midrule
CViT & 0.875 & 0.582 & \textbf{0.321} & 0.880 & 0.603 & 0.343 \\
QSANN \cite{bib30} & 0.845 & 0.509 & 0.284 & 0.850 & 0.485 & 0.317 \\
QSAM \cite{bib40} & 0.875 & 0.534 & 0.290 & 0.867 & 0.530 & 0.341 \\
\midrule
HQViT & \textbf{0.885} & \textbf{0.596} & 0.313 & $\mathbf{0.885}$ & $\mathbf{0.607}$ & $\mathbf{0.345}$ \\
\midrule
\end{tabular}
\label{tab-miniimagenet}
\end{table*}

\subsection{Accuracy performance}\label{subsec5.2}

\textbf{Results on MNIST.} On MNIST, we conduct four binary classification tasks (0 or 1, 1 or 8, odd or even, $>$4 or $\leq4$) and one ten-class classification task (0 to 9). To evaluate the performance of HQViT, we compare it with four other quantum image classifiers, i.e., TTN/MERA \cite{bib38}, two quantum CNN models (marked as QCNN-1 \cite{bib20} and QCNN-2 \cite{bib19}), and a quantum self-attention model (QSAM) \cite{bib40}.

From Table \ref{tab-mnist}, we can see that HQViT achieves the highest accuracy on all classification tasks. For simpler tasks (0/1 classification and 1/8 classification), the prediction accuracy reaches $100\%$. And more impressively, as the task becomes more complex, our model's leading margin becomes larger. For complex binary classification tasks such as ``odd or even" and ``$>$4 or $\leq4$", HQViT achieves accuracy that is about $6-10\%$ higher than the next best models. In the 0-9 multi-class classification task, the accuracy reaches $93.10\%$, surpassing the next best model, QSAM, by nearly $11\%$. This result demonstrates that our model's advantage becomes even more pronounced when tackling more complex tasks. This could be attributed to HQViT’s encoding strategy, which inherently carries positional information. This allows the model to better capture contextual correlations compared to quantum self-attention models that do not incorporate positional encoding.

\textbf{Results on MedMNIST2D.} MedMNIST2D is a large-scale MNIST-like collection of standardized biomedical images, including 12 datasets for 2D images \cite{bib44}. These datasets have image size of $28\times 28$, and cover diverse dataset sizes (from 100 to 100,000) and tasks (binary and multi-class). In the related work, only QOViT \cite{bib42} has conducted experiments on them. Hence, we compare our results with theirs on MedMNIST2D. In Table \ref{tab-medmnist}, we can observe that HQViT outperforms QOViT in 3 out of 5 sub-datasets. Notably, when conducting experiments, QOViT uses classical preprocessing (e.g., fully connected layers) to extract features before encoding them into quantum states \cite{bib42}. By contrast, HQViT uses only simple downsampling or upsampling without any learning-based preprocessing, highlighting the strength of HQViT’s quantum self-attention mechanism in feature extraction.

\textbf{Results on CIFAR-10 and Mini-ImageNet.} CIFAR-10 consists of 10 object categories, while Mini-ImageNet contains 10 main categories, each with 10 subcategories, representing finer distinctions within each category. For each dataset we conduct binary, 4-class, and 10-class prediction tasks. On these datasets, We compare HQViT with QSAM \cite{bib40} and QSANN \cite{bib30}, and also introduce a naive classical Vision Transformer (CViT) baseline, which refers to replacing the quantum self-attention module with its classical equivalent while keeping all other components identical to those in HQViT. The classical equivalent specifically refers to a standard self-attention structure: three fully connected networks of dimension $d\times d$ that map the input data to the $Q$, $K$, and $V$. The attention coefficients are then computed by using Eq. (\ref{attention}), resulting in the weighted $V$. We describe it as 'naive CViT' because its structure is simplified compared to the conventional ViT architecture in \cite{bib29}.

\begin{equation}
\textbf{Attention}(Q, K, V)=\textbf{softmax} (\frac{QK^T}{\sqrt{d}})V \label{attention}
\end{equation}

Firstly, as seen in Tables \ref{tab-cifar10} and \ref{tab-miniimagenet}, HQViT achieves superior performance over QSANN \cite{bib30} and QSAM \cite{bib40} in both two datasets across all tasks. The main difference between HQViT and QSANN \cite{bib30}/\cite{bib40} lies in its direct calculation of the similarity between $Q$ and $K$ in the quantum-enhanced feature space (rather than in the classical feature space). This may allow HQViT to capture more expressive similarity features than its two peers, leading to its improved classification performance. Secondly, HQViT outperforms its classical counterpart, naive CViT, in most tasks, further demonstrating that the feature extraction capability of the quantum-enhanced space is more powerful than that of the classical feature space of the same scale. This highlights the potential of quantum computing to enhance the performance of machine learning models.

\subsection{Effect of Model Configurations}\label{subsec5.3}

\begin{figure}[h]
    \centering
    \includegraphics[width=0.5\textwidth]{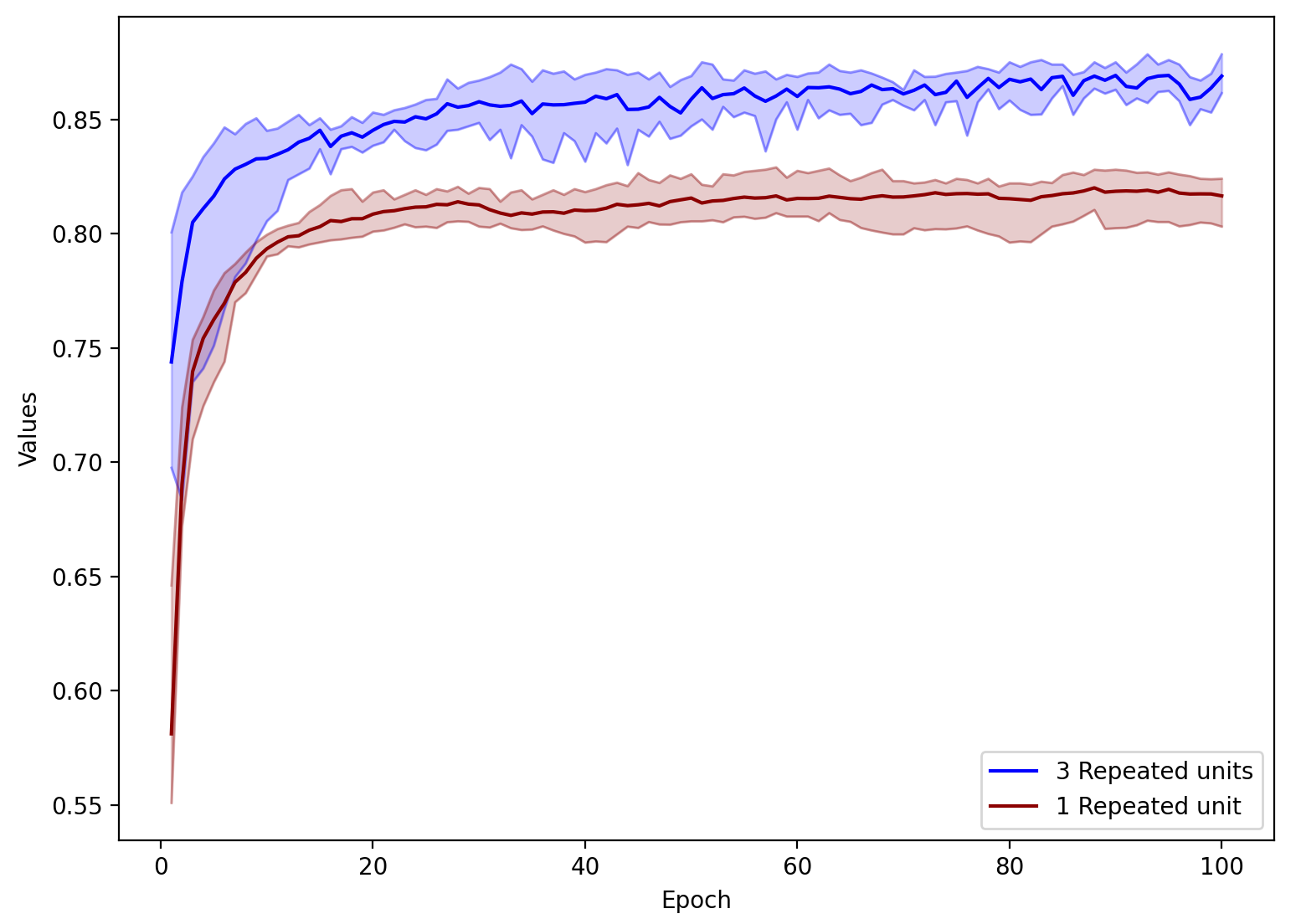}
    \caption{The impact of the number of repeated units in the PQC on performance. The selected scenario in the figure is based on the CIFAR-10 dataset, with image size of 16x16, and 3 QTBs. The bold curves represent the averages of five experiments, with the shaded areas indicating the deviations.} \label{cifar10_1-3unit}
\end{figure}

\begin{figure}[h]
    \centering
    \includegraphics[width=0.5\textwidth]{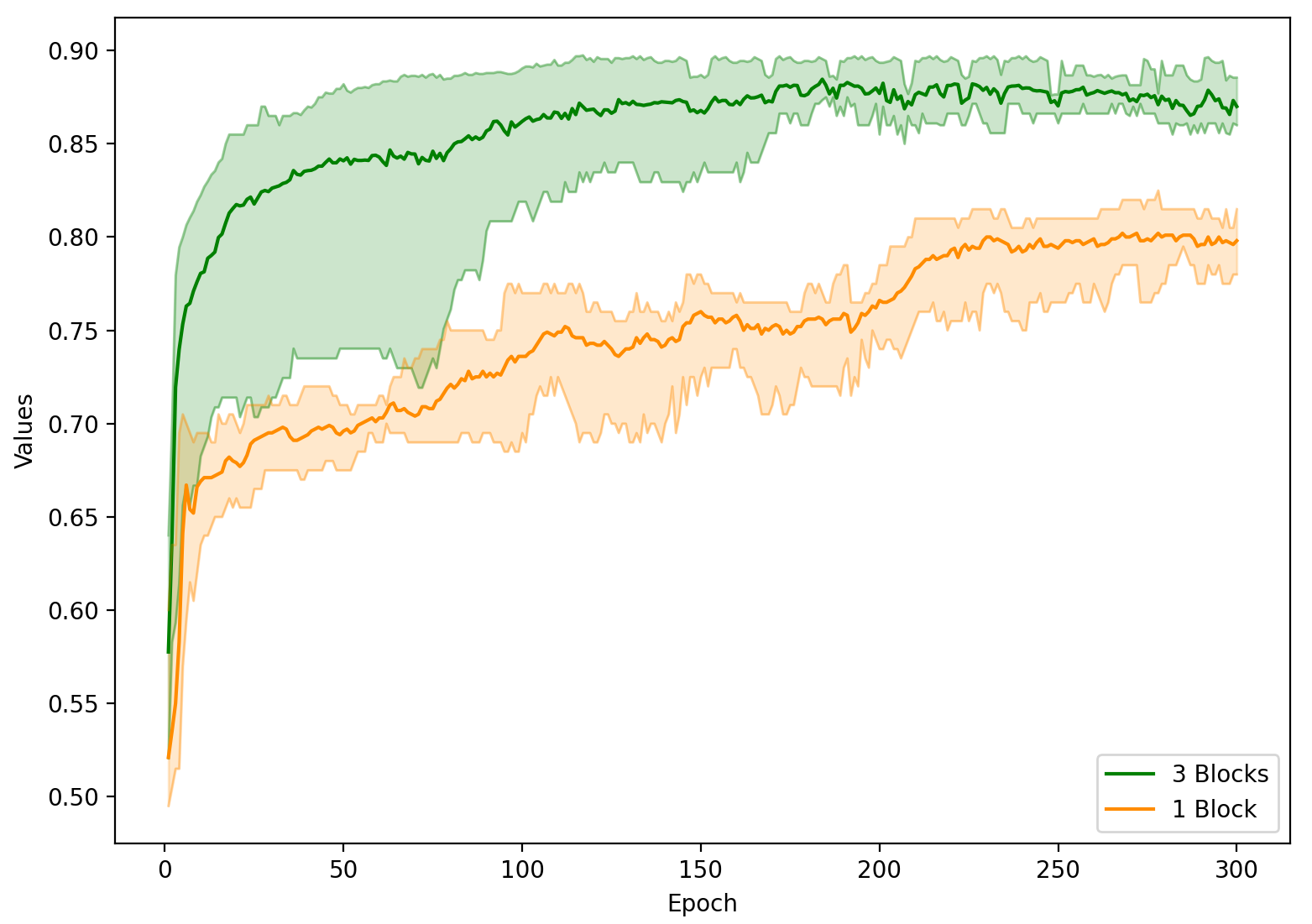}
    \caption{The impact of the number of QTBs on performance. The selected scenario in the figure is based on the Mini-ImageNet dataset, with image size of 32x32, and 3 repeated units in the PQC. The bold curves represent the averages of five experiments, with the shaded areas indicating the deviations.} \label{miniimagenet_1-3block}
\end{figure}

We have investigated the impact of key experimental settings, including the number of repeatable basic units in PQC and the number of QTBs. Both are contributors to the model depth, influencing model's expressibility. As shown in Fig. \ref{cifar10_1-3unit}-\ref{miniimagenet_1-3block}, increasing the number of PQC repeatable units and stacking more QTBs both lead to better performance. This shows that HQViT scales well with depth. However, beyond three units or QTBs, the model’s training speed slows down, and its performance tends to saturate. Therefore, in our experiments, we set the number of QVBs to 3 and the number of repeated units in PQC to 2 or 3.  In addition, we observe that the model successfully converges after several epochs, demonstrating outstanding training behavior.


\section{Conclusion}\label{sec6}

In this paper, we have proposed an efficient hybrid quantum vision transformer (HQViT) for image classification. Our complexity analysis and experiments demonstrate that HQViT significantly reduces classical computational complexity by consuming only limited quantum resources, while delivering excellent classification accuracy. Its performance not only surpasses that of existing quantum transformer models but also outperforms other quantum image processing models, such as QCNN, particularly excelling in complex tasks involving large-size images. This highlights HQViT’s ability to leverage the power of quantum computing for vision tasks, marking a new direction for quantum-enhanced models in the computer vision domain.

Three key factors contribute to our model’s success:
i) Selective quantum-classical hybridization. Rather than pursuing a full quantumization of a vision transformer, which would require an excessive number of ancilla qubits and deep quantum circuits, we selectively retain certain classical components. This ensures that the model remains practical for real-world tasks while maintaining high accuracy.
ii) Whole-image processing. HQViT processes the entire image in one step, allowing the attention coefficient matrix to be derived through the internal evolution of a quantum circuit. This approach simplifies the model structure, better harnesses quantum advantages, and reduces classical computational complexity.
iii) Efficient use of quantum resources. Amplitude encoding enables the model to process the whole image with a relatively small number of qubits, making it feasible for NISQ devices. This encoding strategy also inherently provides positional information at the input stage, improving the model’s contextual understanding without additional positional encoding. However, incorporating explicit positional encoding methods, as seen in classical self-attention, remains a future direction to explore for further enhancing the model’s capabilities.

Our model has been tested on various datasets, well demonstrating its scalability. HQViT’s width can be easily extended by increasing the number of qubits, enabling the model to handle exponentially larger input sizes. On the other hand, the model's depth can be increased by increasing the repeated basic units and stacking more QTBs. However, this requires data conversion between quantum and classical states at the interface of two adjacent QTBs, which adds overhead for encoding and measurement. This may be a necessary trade-off to maintain the model's practicality and scalability in depth. In heavy and fully quantum models, typically only a single transformer block is used, as stacking multiple blocks is not feasible, leading to a poor scalability in depth. Therefore, developing a method to stack multiple QTBs without requiring data conversion between classical and quantum states emerges as a key direction for future research.





\ifCLASSOPTIONcaptionsoff
  \newpage
\fi



%

\bibliographystyle{IEEEtran}
\bibliography{mybib} 

\clearpage 

\begin{appendices}

\renewcommand{\thesection}{Supplementary Materials \Alph{section}} 
\renewcommand{\appendixname}{Supplementary Materials}  

\section{}\label{secA1}

We give the detailed derivation of Eq. (\ref{eq-P(0)}) presented in Section \ref{subsec4.4}.
From Eq. (\ref{eq4.4-1}), we can derive the probability of measuring outcome 00 on the ancilla qubit as:

\begin{align}
Pr(0) & = | \frac{1}{2} \left( |\Psi_2\rangle + \Psi_{sw}\rangle ) \right|^2  \nonumber \\ 
& = \frac{1}{4} ( \langle \Psi_2|\Psi_2 \rangle + \langle \Psi_2|\Psi_{sw}\rangle + \langle \Psi_{sw}|\Psi_2\rangle + \langle \Psi_{sw}|\Psi_{sw}\rangle ) \nonumber \\
& = \frac{1}{2} ( 1 + \langle \Psi_2|\Psi_{sw}\rangle )  \label{eqA-1}
\end{align}
Substituting Eq. (\ref{eq4.4-2})-(\ref{eq4.4-3}) into Eq. (\ref{eqA-1}) yields:

\begin{align}
Pr(0) & = \frac{1}{2} \big( 1 + \sum_{i,l,j,k} \alpha_{il}^* \beta_{jk}^* \langle i|\langle l|\langle j|\langle k|(\sum_{i,l,j,k} \alpha_{ik} \beta_{jl} |i\rangle |l\rangle |j\rangle |k\rangle) \big) \nonumber\\
& = \frac{1}{2} ( 1 + \sum_{i,l,j,k} \alpha_{il}^*\beta_{jk}^*\alpha_{ik}\beta_{jl} \langle i|i\rangle \langle l|l\rangle \langle j|j\rangle \langle k|k\rangle ) \nonumber\\
& = \frac{1}{2} ( 1 + \sum_{i,l,j,k} \alpha_{il}^*\beta_{jl}\beta_{jk}^*\alpha_{ik}) \nonumber\\
& = \frac{1}{2} \big( 1 + \sum_{i} \sum_{j} (\sum_{l} \alpha_{il}^*\beta_{jl} \sum_{k} \beta_{jk}^*\alpha_{ik}) \big) \nonumber\\
& = \frac{1}{2} ( 1 + \sum_{i} \sum_{j} |\langle q_i|k_j\rangle|^2 )
\end{align}


\section{}\label{secA2}

We give the detailed derivation of Eq. (\ref{eq-measurement}) presented in Section \ref{subsec4.4}. When measuring the index subsystems of $R_Q$, $R_Q$ and the ancilla qubit, we use a set of measurement operators $\{M_{i,j}\}_{i,j=0}^{T-1}$, where each $M_{i,j}$ is given as:

\begin{equation}
M_{i,j} = |i\rangle \langle i| \otimes \mathbf{I} \otimes |j\rangle \langle j| \otimes \mathbf{I} \otimes |0\rangle \langle 0|,
\end{equation}
where $\{i, j\}$ is a pair of specific index of $q_i$ and $k_j$, and $\mathbf{I}$ is an identity operator performed on the patch subsystems. According to Eq. (\ref{eq3-meassure}), for each $M_{i,j}$ we have:

\begin{align}
Pr(i,j,0) & = \langle \Psi_3 | M_{i,j}^{\dagger} M_{i,j} | \Psi_3 \rangle  \nonumber\\
& = \frac{1}{4} \big( (\langle \Psi_2| + \langle \Psi_{sw}|) \otimes \langle 0| \big) |i\rangle \langle i| \otimes \mathbf{I} \otimes |j\rangle \langle j| \otimes \mathbf{I} \nonumber\\
& \quad \otimes |0\rangle \langle 0| \big( ( |\Psi_2\rangle + |\Psi_{sw}\rangle) \otimes |0\rangle \big)  \nonumber\\
& = \frac{1}{4}(\sum_{l,k}\alpha_{il}^* \beta_{jk}^*\langle l|\langle k| + \sum_{l,k}\alpha_{ik}^* \beta_{jl}^*\langle l|\langle k|) \nonumber\\
& \quad (\sum_{l,k} \alpha_{il} \beta_{jk} |l\rangle |k\rangle + \sum_{l,k} \alpha_{ik} \beta_{jl} |l\rangle |k\rangle)  \nonumber\\
& = \frac{1}{4}(\sum_{l,k}\alpha_{il}^*\alpha_{il}\beta_{jk}^*\beta_{jk} + \sum_{l,k}\alpha_{il}^*\beta_{jl}\beta_{jk}^*\alpha_{ik} \nonumber\\
& \quad + \sum_{l,k}\alpha_{ik}^*\beta_{jk}\beta_{jl}^*\alpha_{il} + 
\sum_{l,k}\alpha_{ik}^*\alpha_{ik}\beta_{jl}^*\beta_{jl})  \nonumber\\
& = \frac{1}{4}(2\langle q_i|q_i\rangle \langle k_j|k_j\rangle + 2\langle q_i|k_j\rangle \langle q_i|k_j\rangle)  \nonumber\\
& = \frac{1}{2}(|q_i|^2|k_j|^2 + |\langle q_i|k_j\rangle|^2).
\end{align}

\end{appendices}

\end{document}